\documentclass[journal]{IEEEtran}

\usepackage{etoolbox}
\usepackage{graphicx}
\usepackage{subcaption}
\usepackage{hyperref}
\usepackage{placeins}
\begin{document}
\title{A ripple in time: a discontinuity \\ in American history}
\author{Alexander Kolpakov$^1$\thanks{${}^1$CSTEM, University of Austin, Texas, USA} and Igor Rivin$^2$\thanks{${}^2$Department of Mathematics, Temple University, USA}}


\maketitle
\begin{abstract}
In this technical note we suggest a novel approach to discover temporal (related and unrelated to language dilation) and personality (authorship attribution) aspects in historical datasets. We exemplify our approach on the State of the Union addresses given by the past 42 US presidents: this dataset is known for its relatively small amount of data, and high variability of the size and style of texts. Nevertheless, we manage to achieve about 95\% accuracy on the authorship attribution task, and pin down the date of writing to a single presidential term.  
\end{abstract}




\section{Results and novelty}\label{s:extended-abstract}

In the present work we use the State of the Union Address (SOTU) dataset from \href{https://www.kaggle.com/datasets/rtatman/state-of-the-union-corpus-1989-2017}{Kaggle} to make some surprising (and some expected) observations pertaining to the general timeline of American history, and the character and nature of the addresses themselves. 

Our main approach is using vector embeddings from transformers, such as BERT (DistilBERT) \cite{devlin2019bert, sanh2020distilbert} and GPT--2 \cite{radford2019language} in conjunction with dimension reduction techniques such as UMAP \cite{mcinnes2018umap}, TriMAP \cite{amid2019trimap} and PaCMAP \cite{tuncer2015pacmap}. 

While it is widely believed that BERT (and its variations) is most suitable for NLP classification tasks, we find out that GPT--2 in conjunction with nonlinear dimension reduction methods, such as UMAP, provides stronger clustering. This makes GPT--2 + UMAP an interesting alternative. In our case, no model fine--tuning is required, and the pre--trained out--of--the--box GPT--2 model is enough. 

We also used a fine--tuned DistilBERT model for classification detecting which President delivered which address, with very good results (accuracy 93\% -- 95\% depending on the run). An analogous task was performed to determine the year of writing, and we were able to pin it down to about 4 years (which is a single presidential term). 

Previous attempts at the authorship attribution tasks used much larger datasets  with a much smaller amount of authors \cite{stanisz2019linguistic}, or did not reach better accuracy except on a smaller ``sure--fire'' subset of the test data \cite{savoy}.

Other related studies \cite{programming-historian, tmgpt} concentrated on fine--tuning LLMs to handle temporal aspects of language (e.g. linking the names of country leaders to a given historical period) without an attempt to discover any relationships in the training data. Even though certain observations can likely be made by prompting the model, the latent space of the transformer was not analysed. 

It is worth noting that SOTU addresses provide relatively small writing samples (with about 8000 words on average, and varying widely from under 2000 words to more than 20000), and that the number of authors is relatively large (we used SOTU addresses of 42 US presidents). This shows that the techniques employed turn out to be rather efficient, while all the computations described in this note can be performed using a single GPU instance of Google Colab.

The accompanying code is available on GitHub \cite{repo} for the purpose of full reproducibility of our study. 

\section{Problems and methods}\label{s:introduction}
 
\subsection{Dataset} The dataset is the collection of complete texts of State of the Union addresses (SOTU) of US presidents, from John Adams to Donald Trump \cite{Kaggle}. We modified the dataset by splitting the addresses into train and test files, such that each President has at least one train SOTU, and at least one test SOTU file (we use the same file for training and testing for Taylor, as it seems that splitting it into, say, halves makes even less sense). Also, some presidents are more represented than others in the SOTU corpus because they spent more years in office. On average, we keep the train / test split about 75\% / 25\% for each individual President.

\subsection{Methods of study} We used BERT and GPT--2 to embed SOTU addresses as high--dimensional vectors. We then used several non--linear dimension reduction techniques (namely, UMAP, TriMAP, and PaCMAP) to obtain their low--dimensional embeddings. All methods produced results that are relatively close to each other. We opine that the following methods did the best clustering job:
\begin{itemize}
    \item GPT--2 embedding followed by UMAP (sharp break at 1928--1929),
    \item DistilBERT embedding followed by TriMAP (sharp brake at 1920--1921). 
\end{itemize}

To complete this investigation, we fine-tuned the DistilBERT model to detect which President delivered which address. The methods and results are described in Section \ref{attribution}. We also used DistilBERT for embeddings instead of the original BERT as it is much faster and lightweight. Thus, our experiments can be easily replicated in a Google Colab notebook within reasonable time. 

\section{Results and observations}\label{s:section}

\subsection{Temporal clustering}

The first thing we saw was largely unsurprising: addresses delivered by the same President were embedded close together, and addresses chronologically close together were also geometrically close together (where we measure "closeness" by the cosine distance between the vectors).  This seems a natural consequence of temporal language dilation. 

\begin{figure}
    \centering
    \begin{subfigure}{0.45\textwidth}
        \centering
        \includegraphics[scale=0.39]{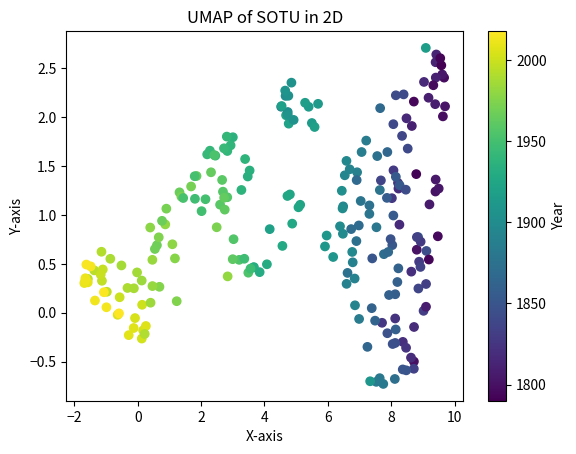}
        \caption{2D UMAP applied to the GPT--2 embedding of SOTU}
        \label{fig:gpt2umap2d_}
    \end{subfigure}
    \hfill
    \begin{subfigure}{0.45\textwidth}
        \centering
        \includegraphics[scale=0.45]{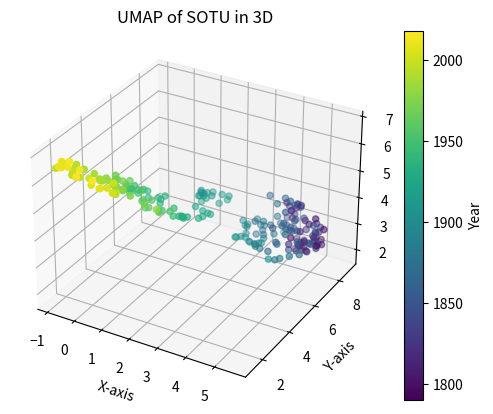}
        \caption{3D UMAP applied to the GPT--2 embedding of SOTU}
        \label{fig:gpt2umap3d_}
    \end{subfigure}
    \caption{UMAP visualizations of the GPT--2 embedding of SOTU}
    \label{fig:gpt2umap}
\end{figure}

\begin{figure}
    \centering
    \begin{subfigure}{0.45\textwidth}
        \centering
        \includegraphics[scale=0.4]{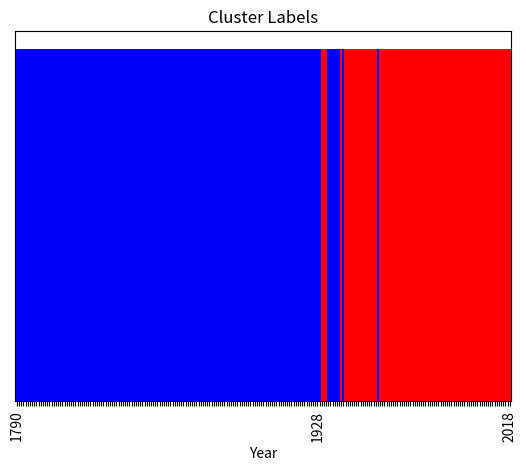}
        \caption{Temporal clustering of the GPT--2 embedding of SOTU: separation starts around 1941}
        \label{fig:gpt2cluster}
    \end{subfigure}
    \hfill
    \begin{subfigure}{0.45\textwidth}
        \centering
        \includegraphics[scale=0.4]{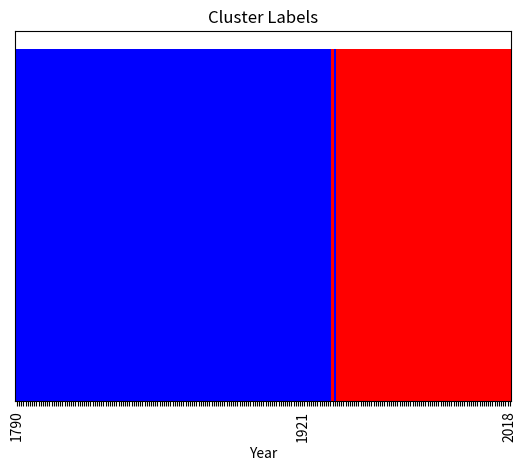}
        \caption{Temporal clustering of the DistilBERT embedding of SOTU: separation starts around 1937}
        \label{fig:bert2cluster}
    \end{subfigure}
    \caption{Temporal clustering of SOTU embeddings}
    \label{fig:embeddingclusters}
\end{figure}

What was surprising, however, is a large break, as demonstrated by the UMAP visualization (Figures \ref{fig:gpt2umap2d_} -- \ref{fig:gpt2umap3d_}), between the addresses written before 1927 and those written after 1932. We were rather curious to see this, and did cluster analysis on the original vector embedding (without reducing it by UMAP), and the results were confirmed (with a couple of minor differences) -- see Figure~\ref{fig:embeddingclusters}.

\subsection{BERT vs GPT--2}

Even though BERT is widely assumed to provide the right kind of embedding for NLP classification tasks, we can clearly see that GPT--2 in conjunction with UMAP produces very good clustering, too. However, the clustering in the initial DistilBERT embedding has a much clearer temporal separation than in the case of GPT--2 (compare Figure~\ref{fig:gpt2cluster} to Figure~\ref{fig:bert2cluster}). 

Usually, vector embedding comes as the initial part of a model, and thus the most important factor is how much information it can pass to the subsequent layers, or methods. In this case we posit that GPT--2 managed to capture enough information in order for UMAP to produce notably better clustering than DistilBERT would produce alone. Since GPT--2 is a generative model, we can also speculate that its embedding captures more granular details of the dataset, in general. This additional granularity is not obvious, and can only be exposed by using additional features such as non--linear dimension reduction.  

\section{Authorship attribution of the addresses}
\label{attribution}

We have also experimented with determining which President wrote which State of the Union address. The simplest solution was the one based on pre--trained embedding (just as was done before), and this had an accuracy of around 80\%, which, while not terrible, was not very exciting. So, as the next step, we fine--tuned a HuggingFace DistilBERT model for the classification task. 

Since the DistilBERT model can only deal with 512 tokens (shorter than any of our addresses), we chunked each texts with a sliding window of 512 token having an overlap of 128 tokens, and treated each chunk as a training sample. For the testing phase, we did the same, but aggregated the scores for the chunks to get a prediction for the document. 

It should be noted that the per--chunk prediction rate was not great -- something south of 60\%. Not too surprisingly, the wrong answers of the model seem to be for temporally adjacent authors (Washington and Adams, or Carter and Ford), so presumably the Zeitgeist and language matter more than the actual politics.

However, instead of considering individual chunks of text, a better way to capture its context and vernacular is to aggregate the logits output by the model over all chunks produced from that text. After that, the argmax is applied. In this case the accuracy raises spectacularly up to $\approx$ 93\% -- 95\%, depending on the number of fine--tuning epochs. 

We should note here that the amount of training data for each author is relatively small, and also varies greatly: from 1790 words in an average SOTU address for John Adams to 22614 words on average for William H. Taft \cite{peters1999length}. The average length of a SOTU address is just about 8358 words. 

For comparison, the study carried out in \cite{stanisz2019linguistic} used novels for authorship attribution. The average word length of a novel is believed to be over 40000. The number of authors for authorship attribution in \cite{stanisz2019linguistic} is only 8, with classification accuracy of $\approx$ 85\%--90\%. 

Another, less recent study \cite{savoy} that targets SOTU addresses specifically, produces a comparably low accuracy. By the author's own admission, it has a higher success rate of 95.7\% as applied to an artificially created category based off similarity distances. The actual accuracy of the approach described in \cite{savoy} can be computed as $ (45+66+79) / (45+71+108) \approx 84.8 \%$ (``Authorship attribution'', page 1652).

In the present work, we manage to identify all 42 authors with at least 93\% accuracy in an honest--to--goodness test run, and with training based on a much smaller amount of data than 6 novels written by each. 

\section{Year of writing of an address}
\label{year}

We also attempted to determine the year in which SOTU addresses were written. Here we used a customization of DistilBERT with several dense layers having GELU activation. While training the resulting model for the regression task, we suggest using the $L_p$ loss norm with $p \geq 3$, to reduce the influence of possible outliers in the temporal distribution of SOTU addresses.

Indeed, we conclude that, among the values of $p \in [1,5]$, having $p=3$ gives us the best results. In fact, $p$ does not have to be an integer: we did not do full--scale hyperparameter optimization for the purpose of this study. The model architecture and the best value of $p$ were found by trial and error from a relatively small search space amenable to manual inspection. 

As in the case of authorship attribution, the model's accuracy on chunks of text is quite poor: we get about $\approx 7.3$ years RMSE. 

The best RMSE is $\approx 4.5$ years, which is much closer to one presidential term. This is achieved by averaging all year predictions over text chunks. We would agree that this result is not spectacular, however not bad either given that the language and policy may not change much over such a short period of time. 

\section{Discussion}

We will attempt to explain our results in the historical context, and make several conjectures about the root causes.  

The first one is that before the time of Franklin Roosevelt, it was extremely uncommon for presidents to use speechwriters, while Roosevelt used their help a lot. It seems plausible that an address written by a hired hand will be different from one written by the President of the United States (even though US presidents obviously retain considerable editorial control to the present day).

The second is that the nature of the United States changed considerably after World War II. Before it was a very large, very rich, but remote and provincial country, and after it became a world dominating empire, with the attendant difference in focus, and hence a considerable difference in the emphasis of presidential addresses. Likely the first changes have already appeared earlier, which explains that the ripple happens sometime in the 1920's. 

The colors in, say, the UMAP embedding of SOTU (Figures \ref{fig:gpt2umap2d_} and \ref{fig:gpt2umap3d_}) change continuously representing the language shift. However, the break is sharp, and thus likely happens for a different reason than just language dilation in time. 

We would be happy to hear other theories and possible explanations. What is obvious is that a considerable shift in SOTU is detectable by a variety of methods, and thus there must be an underlying reason (or even a multitude of reasons). 


\section{Data accessibility} All numerical experiments can be reproduced by accessing the associated GitHub repository \cite{repo}.


\section{Authors’ contributions} A.K. and I.R. conceived the mathematical models, interpreted the computational results, and composed the manuscript. A.K. and I.R. implemented and performed all the numerical experiments, and otherwise contributed in equal parts. 

\section{Acknowledgments} This work was supported by the Google~Cloud Research Award number GCP19980904.

\bibliography{bibliography}

\appendix
\section{Dimension reductions of GPT--2 embedding}

\begin{figure}[h]
    \centering
    \begin{subfigure}{0.32\textwidth}
        \centering
        \includegraphics[scale=0.3]{figures/sotu_gpt2_umap_2d.png}
        \caption{2D UMAP}
        \label{fig:gpt2umap2d}
    \end{subfigure}
    \hfill
    \begin{subfigure}{0.32\textwidth}
        \centering
        \includegraphics[scale=0.3]{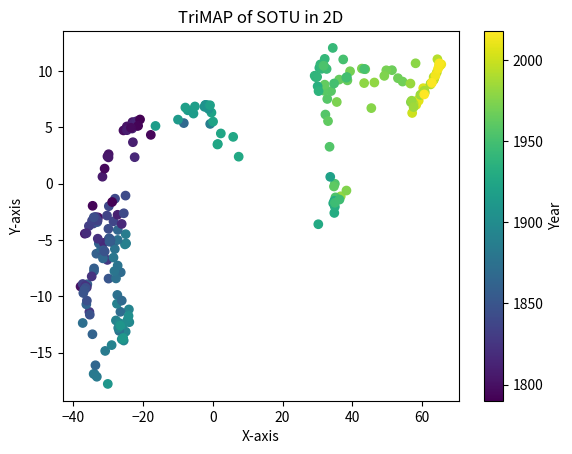}
        \caption{2D TriMAP}
        \label{fig:gpt2trimap2d}
    \end{subfigure}
    \hfill
    \begin{subfigure}{0.32\textwidth}
        \centering
        \includegraphics[scale=0.3]{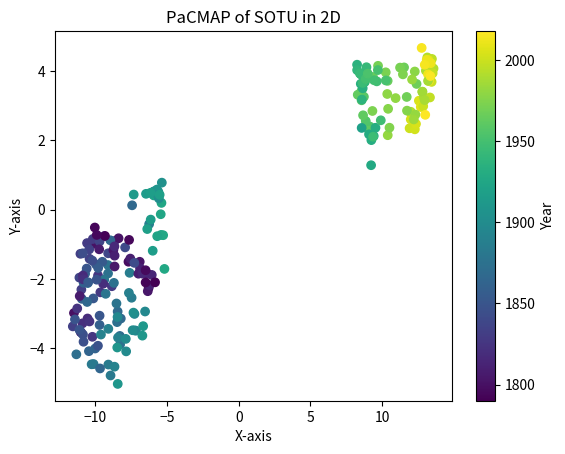}
        \caption{2D PaCMAP}
        \label{fig:gpt2pacmap2d}
    \end{subfigure}
    \caption{2D visualizations of the GPT--2 embedding of SOTU}
    \label{fig:gpt2reduction2}
\end{figure}

\begin{figure}[h]
    \centering
    \begin{subfigure}{0.32\textwidth}
        \centering
        \includegraphics[scale=0.35]{figures/sotu_gpt2_umap_3d.png}
        \caption{3D UMAP}
        \label{fig:gpt2umap3d}
    \end{subfigure}
    \hfill
    \begin{subfigure}{0.32\textwidth}
        \centering
        \includegraphics[scale=0.35]{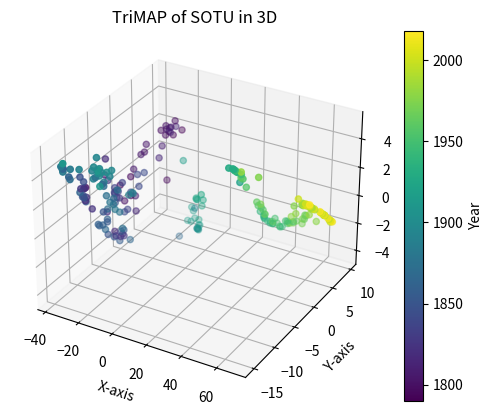}
        \caption{3D TriMAP}
        \label{fig:gpt2trimap3d}
    \end{subfigure}
    \hfill
    \begin{subfigure}{0.32\textwidth}
        \centering
        \includegraphics[scale=0.35]{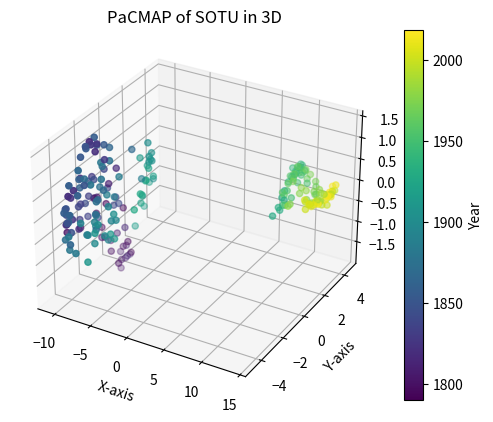}
        \caption{3D PaCMAP}
        \label{fig:gpt2pacmap3d}
    \end{subfigure}
    \caption{3D visualizations of the GPT--2 embedding of SOTU}
    \label{fig:gpt2reduction3}
\end{figure}

Figure~\ref{fig:gpt2reduction2} shows the GPT--2 embedding of SOTU reduced to dimension 2 with UMAP\cite{mcinnes2018umap}, TriMAP\cite{amid2019trimap} and PaCMAP\cite{tuncer2015pacmap}, respectively. Figure~\ref{fig:gpt2reduction3} shows analogous reductions to dimension 3 instead. 

\begin{figure}[h]
    \centering
    \begin{subfigure}{0.32\textwidth}
        \centering
        \includegraphics[scale=0.33]{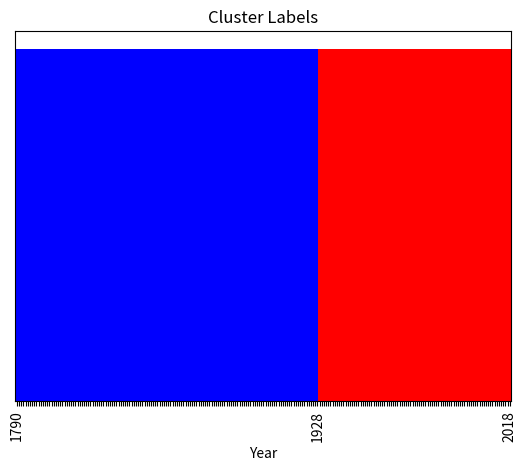}
        \caption{UMAP}
        \label{fig:umap3d_temporal}
    \end{subfigure}
    \hfill
    \begin{subfigure}{0.32\textwidth}
        \centering
        \includegraphics[scale=0.33]{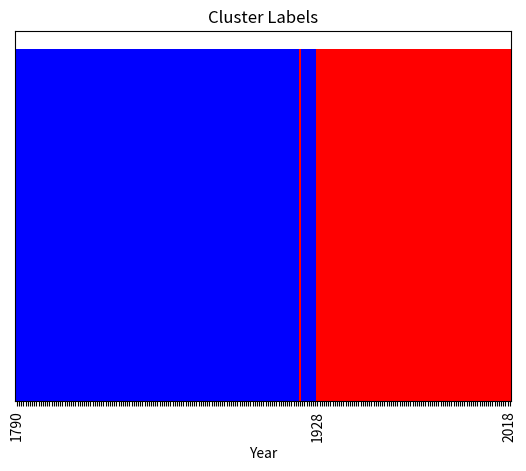}
        \caption{TriMAP}
        \label{fig:trimap3d_temporal}
    \end{subfigure}
    \hfill
    \begin{subfigure}{0.32\textwidth}
        \centering
        \includegraphics[scale=0.33]{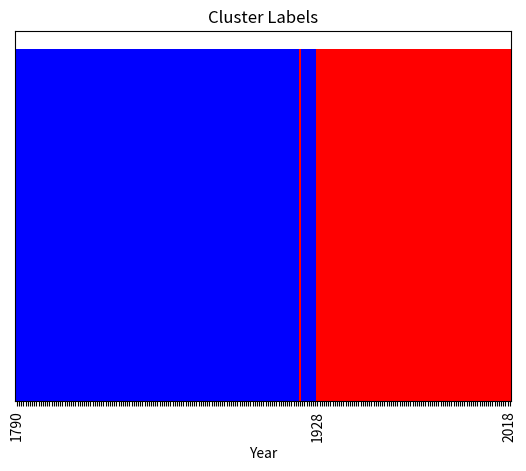}
        \caption{PaCMAP}
        \label{fig:pacmap3d_temporal}
    \end{subfigure}
    \caption{Temporal clustering of SOTU addresses: GPT--2 embedding followed by a dimension reduction technique}
    \label{fig:gpt2temporal}
\end{figure}

The most important part is the temporal aspect of SOTU addresses, not their clustering per se. This is shown in Figure~\ref{fig:gpt2temporal}, where we see that the clearest separation belongs to UMAP. However, TriMAP and PaCMAP are not much worse: in fact, the three cases turn out to be quite comparable. 

In all pictures, temporal separation happens at the break of 1927 and 1928. For the original GPT--2 embedding it is not quite the case (compare to Figure~\ref{fig:gpt2cluster}). 

\section{Dimension reductions of DistilBERT embedding}

\begin{figure}[h]
    \centering
    \begin{subfigure}{0.32\textwidth}
        \centering
        \includegraphics[scale=0.33]{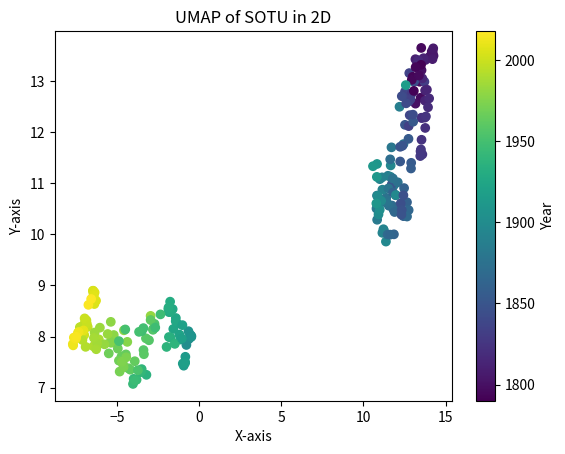}
        \caption{2D UMAP}
        \label{fig:bert_umap2d}
    \end{subfigure}
    \hfill
    \begin{subfigure}{0.32\textwidth}
        \centering
        \includegraphics[scale=0.33]{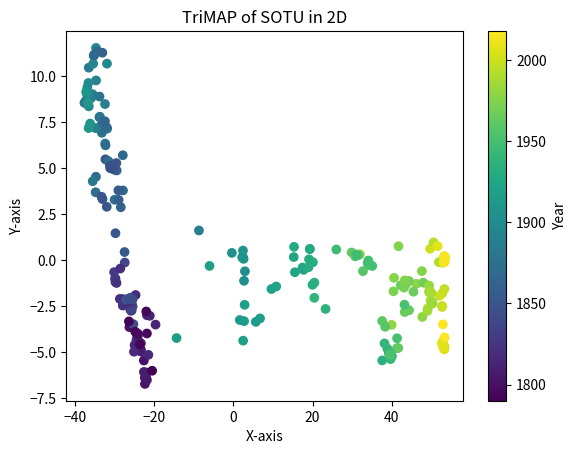}
        \caption{2D TriMAP}
        \label{fig:bert_trimap2d}
    \end{subfigure}
    \hfill
    \begin{subfigure}{0.32\textwidth}
        \centering
        \includegraphics[scale=0.33]{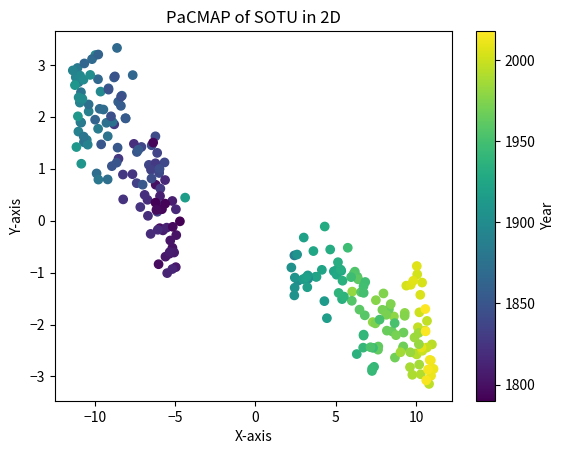}
        \caption{2D PaCMAP}
        \label{fig:bert_pacmap2d}
    \end{subfigure}
    \caption{2D visualizations of the DistilBERT embedding of SOTU}
    \label{fig:bert_reduction2}
\end{figure}

\begin{figure}[h]
    \centering
    \begin{subfigure}{0.32\textwidth}
        \centering
        \includegraphics[scale=0.35]{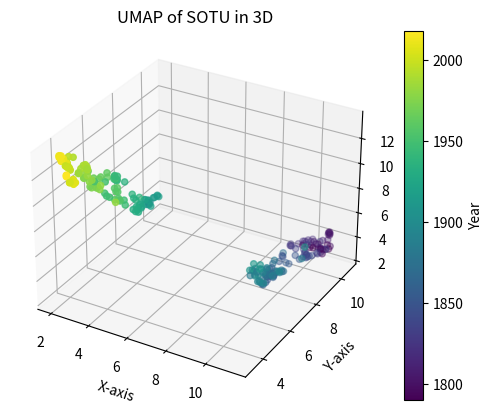}
        \caption{3D UMAP}
        \label{fig:bert_umap3d}
    \end{subfigure}
    \hfill
    \begin{subfigure}{0.32\textwidth}
        \centering
        \includegraphics[scale=0.35]{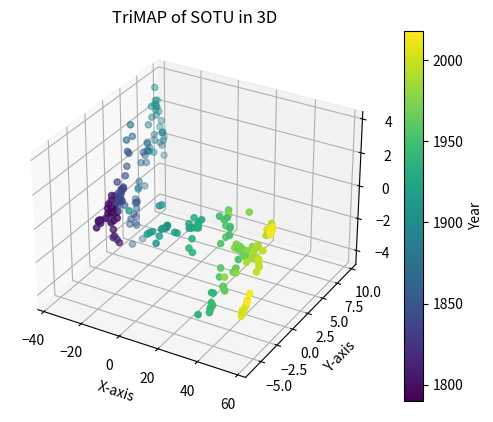}
        \caption{3D TriMAP}
        \label{fig:bert_trimap3d}
    \end{subfigure}
    \hfill
    \begin{subfigure}{0.32\textwidth}
        \centering
        \includegraphics[scale=0.35]{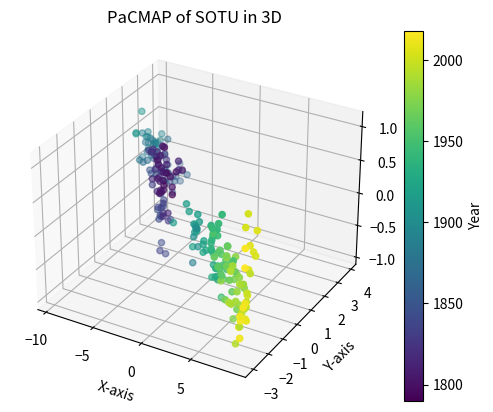}
        \caption{3D PaCMAP}
        \label{fig:bert_pacmap3d}
    \end{subfigure}
    \caption{3D visualizations of the DistilBERT embedding of SOTU}
    \label{fig:bert_reduction3}
\end{figure}

Figure~\ref{fig:bert_reduction2} shows the GPT--2 embedding of SOTU reduced to dimension 2 with UMAP, TriMAP and PaCMAP, respectively. Figure~\ref{fig:bert_reduction3} shows analogous reductions to dimension 3. 

Again, we are interested in the temporal aspect of clustering that is illustrated in Figure~\ref{fig:gpt2temporal}. There, the clearest separation belongs to TriMAP. 

\begin{figure}[h]
    \centering
    \begin{subfigure}{0.32\textwidth}
        \centering
        \includegraphics[scale=0.33]{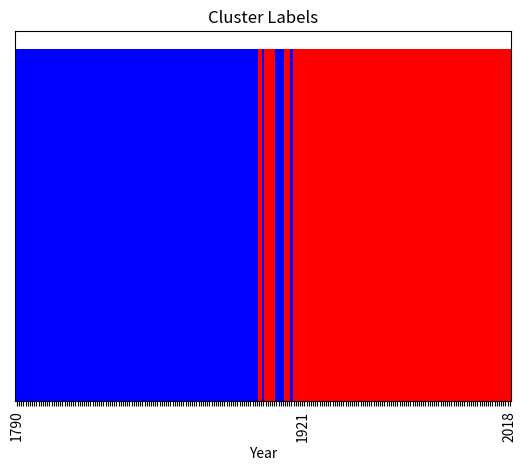}
        \caption{UMAP}
        \label{fig:umap3d_temporal_}
    \end{subfigure}
    \hfill
    \begin{subfigure}{0.32\textwidth}
        \centering
        \includegraphics[scale=0.33]{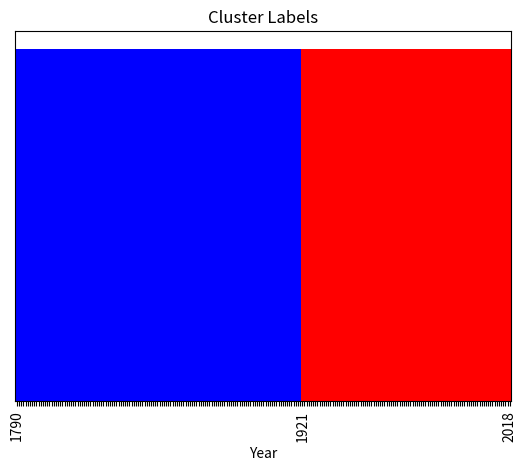}
        \caption{TriMAP}
        \label{fig:trimap3d_temporal_}
    \end{subfigure}
    \hfill
    \begin{subfigure}{0.32\textwidth}
        \centering
        \includegraphics[scale=0.33]{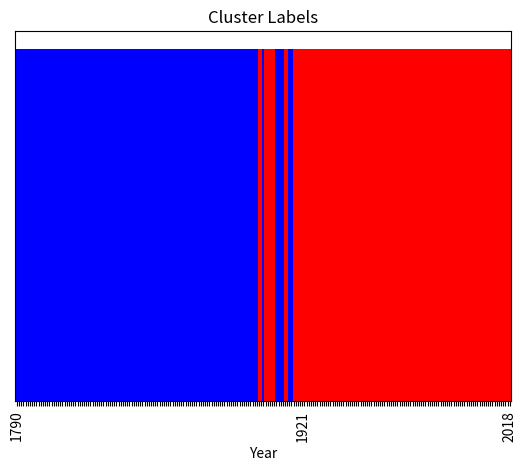}
        \caption{PaCMAP}
        \label{fig:pacmap3d_temporal_}
    \end{subfigure}
    \caption{Temporal clustering of SOTU addresses: DistilBERT embedding followed by a dimension reduction technique}
    \label{fig:dbert2temporal}
\end{figure}

In all pictures, temporal separation happens at the break of 1920 and 1921, though only one of the pictures shows a clear watershed moment. For the original DistilBERT embedding it is not quite the case: the temporal separation in Figure~\ref{fig:bert2cluster} is quite clear, although it happens about 16 years after 1921.  

\FloatBarrier

\end{document}